\documentclass[conference]{IEEEtran}
\IEEEoverridecommandlockouts
\usepackage{cite}
\usepackage{amsmath,amssymb,amsfonts}
\usepackage[table]{xcolor} 
\usepackage[colorlinks=true, linkcolor=blue, urlcolor=magenta, citecolor=green] {hyperref}
\usepackage{algorithmic}
\usepackage{booktabs}
\usepackage{threeparttable} 
\usepackage{hyperref} 
\usepackage{graphicx}
\usepackage{textcomp}
\usepackage{xcolor}
\def\BibTeX{{\rm B\kern-.05em{\sc i\kern-.025em b}\kern-.08em
    T\kern-.1667em\lower.7ex\hbox{E}\kern-.125emX}}
\begin{document}

\title{Geometry-Consistent 4D Gaussian Splatting for
Sparse-Input Dynamic View Synthesis\\
\thanks{\textsuperscript{$*$}Corresponding Author: Yinfeng Cao and Divya Saxena.}
}

\author{\IEEEauthorblockN{Yiwei Li\textsuperscript{$\dagger$}, Jiannong Cao\textsuperscript{$\dagger$}, Penghui Ruan\textsuperscript{$\dagger$}\textsuperscript{$\ddagger$}, Divya Saxena\textsuperscript{$*$}\textsuperscript{$\S$}, Songye Zhu\textsuperscript{$\dagger$}, Yinfeng Cao\textsuperscript{$*$}\textsuperscript{$\dagger$}\textsuperscript{$\sharp$}}
\IEEEauthorblockA{\textit{
\textsuperscript{$\dagger$}The Hong Kong Polytechnic University, Hong Kong, China}\\
\textit{\textsuperscript{$\ddagger$}Southern University of Science and Technology, Shenzhen, China}\\
\textit{\textsuperscript{$\S$}Indian Institute of Technology, Jodhpur, India} \\
\textit{\textsuperscript{$\sharp$}Institute of Cyberspace Technology, HKCT Institute of Higher Education, Hong Kong, China}
\\ 
\{yi-wei.li, penghui.ruan\}@connect.polyu.hk, csjcao@comp.polyu.edu.hk, \\ divyasaxena@iitj.ac.in, ceszhu@polyu.edu.hk, kevincao@ctihe.edu.hk}
}

\maketitle

\begin{abstract}

Gaussian Splatting has been considered as a novel way for view synthesis of dynamic scenes, which shows great potential in AIoT applications such as digital twins. However, recent dynamic Gaussian Splatting methods significantly degrade when only sparse input views are available, limiting their applicability in practice. The issue arises from the incoherent learning of 4D geometry as input views decrease. This paper presents GC-4DGS, a novel framework that infuses geometric consistency into 4D Gaussian Splatting (4DGS), offering real-time and high-quality dynamic scene rendering from sparse input views. While learning-based Multi-View Stereo (MVS) and monocular depth estimators (MDEs) provide geometry priors, directly integrating these with 4DGS yields suboptimal results due to the ill-posed nature of sparse-input 4D geometric optimization. To address these problems, we introduce a dynamic consistency checking strategy to reduce estimation uncertainties of MVS across spacetime. Furthermore, we propose a global-local depth regularization approach to distill spatiotemporal-consistent geometric information from monocular depths, thereby enhancing the coherent geometry and appearance learning within the 4D volume. Extensive experiments on the popular N3DV and Technicolor datasets validate the effectiveness of GC-4DGS in rendering quality without sacrificing efficiency. Notably, our method outperforms RF-DeRF, the latest dynamic radiance field tailored for sparse-input dynamic view synthesis, and the original 4DGS by 2.62dB and 1.58dB in PSNR, respectively, with seamless deployability on resource-constrained IoT edge devices. Our code is available at \href{https://github.com/liyw420/GC-4DGS}{https://github.com/liyw420/GC-4DGS}.

\end{abstract}

\begin{IEEEkeywords}
dynamic view synthesis, 4D Gaussian Splatting, sparse input views, geometric consistency
\end{IEEEkeywords}

\section{Introduction}

Dynamic view synthesis (DVS) aims to model dynamic scenes from videos and render photorealistic novel views for immersive visualizations, which is crucial for AIoT applications such as telepresence \cite{adil2024role} and digital twin creation \cite{cao2024polytwin, cao2025decentralized, cao2025chaintwin}. Recently, this task has witnessed great improvements with the emergence of Gaussian Splatting \cite{kerbl20233d}, an explicit point-based radiance field using a fast differentiable rendering pipeline. However, most dynamic Gaussian Splatting methods \cite{yang2024real,li2024spacetime, wu20244d, bae2025per} require multi-view data captured by a dense array of synchronized cameras (typically 12-50 cameras) to achieve high-fidelity rendering, which restricts their real-world applicability \cite{somraj2024factorized}. For instance, in smart transportation systems, cost and privacy constraints \cite{yang2022coalition, lin2025quality} often restrict cameras to only a few fixed locations, resulting in sparse and limited viewpoints for capturing dynamic traffic flows \cite{ukyo2025robust}. Furthermore, bandwidth constraints and packet loss in wireless and IoT networks \cite{wu2025joint, imran2024exploring} can degrade or interrupt video streams, reducing the number of available viewpoints. Given these prevalent issues in device availability and video transmission within AIoT systems, it is crucial to develop Gaussian Splatting models capable of learning dynamics from sparse views. 

\begin{figure}[t]
  \centering
   \includegraphics[width=1.0\linewidth]{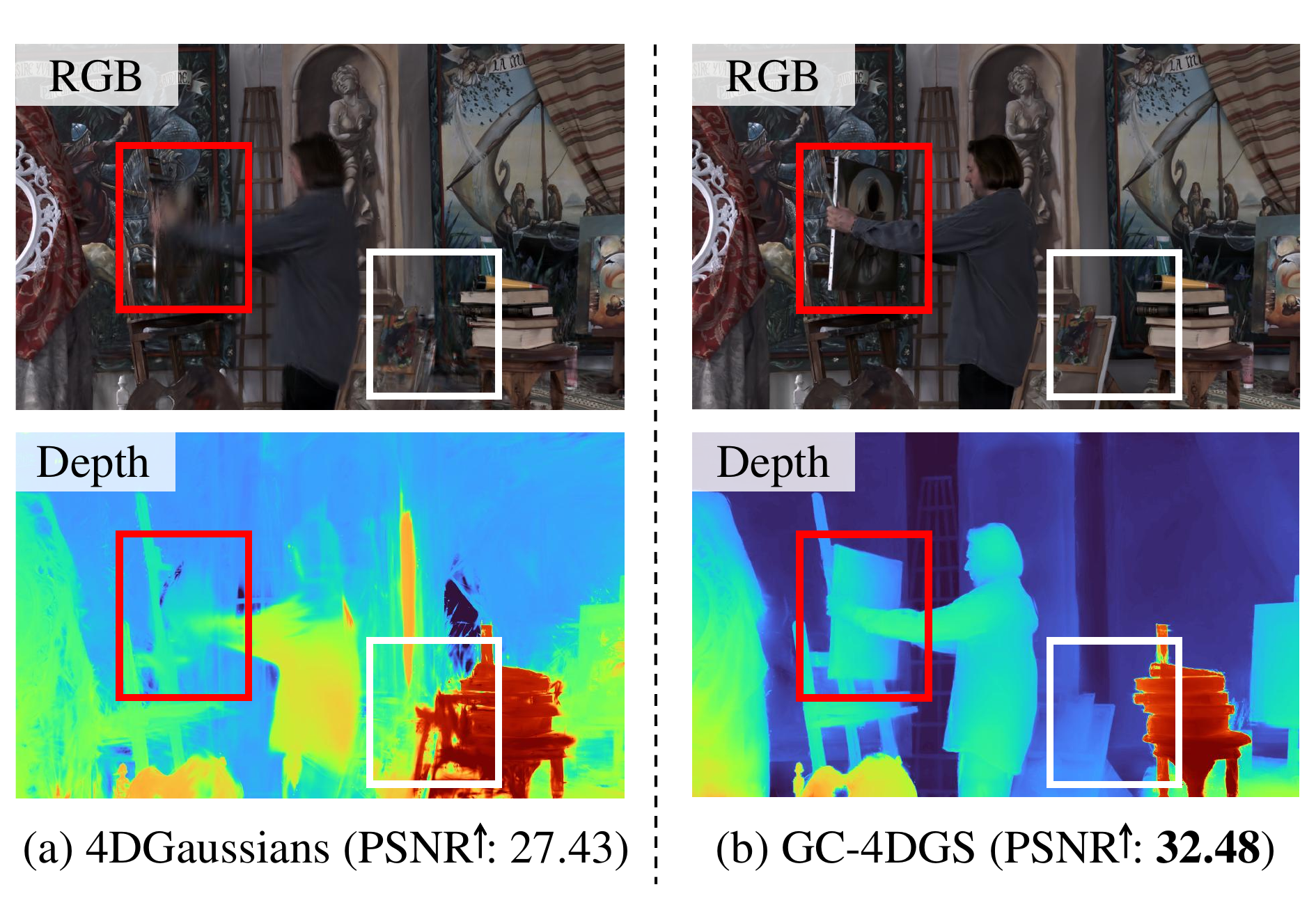}
   \caption{Our Geometry-Consistent 4D Gaussian Splatting (GC-4DGS) achieves high-fidelity rendering quality with only 3 input views. (a) Existing dynamic Gaussian Splatting methods, e.g., 4DGaussians \protect\cite{wu20244d}, learn incorrect 4D geometry from sparse training views. (b) GC-4DGS solves this issue by learning consistent geometry from both MVS and monocular depths, achieving realistic appearance and coherent geometry of dynamic scenes.}
   \label{fig.1}
\end{figure}

Optimizing Gaussian radiance fields with sparse inputs is challenging due to the under-constrained nature of the problem. The unstructured point-based representations struggle to initialize from the extremely sparse point clouds estimated by the COLMAP pipeline\cite{schoenberger2016sfm, schoenberger2016mvs}. Although the learning-based MVS \cite{cao2022mvsformer} provides dense 3D points, directly combining it with dynamic Gaussians leads to suboptimal results due to redundant point estimation and accumulated geometric matching errors. Another line of research \cite{li2024dngaussian,zhu2025fsgs,paliwal2025coherentgs} uses prior depth information from pre-trained MDEs to guide Gaussian optimization. While these methods effectively mitigate geometry degradation for static Gaussians, they overlook the issue of temporal inconsistency in the estimated depth across frames \cite{qingming2025modgs}. Specifically, MDE-derived depth maps inherently lack scale consistency \cite{wang2023sparsenerf}, i.e., the depth ratios between the same pixel pairs can vary across frames, even after normalization. Consequently, simply applying these methods makes it difficult to render high-quality dynamic scenes for time-dependent Gaussian Splatting with sparse inputs.

To address the aforementioned challenges, we introduce Geometry-Consistent 4D Gaussian Splatting (GC-4DGS), a novel DVS framework to achieve real-time high-fidelity dynamic scene rendering from sparse input views. We build our work on 4DGS \cite{yang2024real}, a powerful and coherent spatiotemporal representation to model dynamic scenes. To reduce estimation uncertainties in MVS, a dynamic consistency checking strategy is introduced to dynamically aggregate the depth pixels with low matching errors across spacetime. Specifically, given reprojection errors, neighbor views, and timestamps, we treat dynamic geometric matching cost as consistency among all views and nearby frames, thereby achieving more robust metric depth estimation, which is used for subsequent depth fusion and 4D structure supervision. 

% To tackle the temporal inconsistency in the MDE-derived depth maps, we propose a global-local depth regularization method. We observed that while depth scales may differ across frames, the relative depth order between pixel pairs should remain consistent. This insight led us to introduce a global ranking loss that distills temporal-consistent depth relationships to guide 4DGS learning. At the local scale, we implement a patch-based smoothing constraint to address piece-wise discontinuities in the local geometries. Moreover, we found that 4DGS is very prone to densification failures under sparse view settings, and an adaptive Step Decay strategy is proposed to improve optimization stability. Our approach guides the learning process by promoting global consistency across spacetime while minimizing localized artifacts and prompting stable optimization, enhancing the coherent geometry and appearance learning within the 4D volume.

To tackle the spatiotemporal inconsistency in the MDE-derived depth maps, we propose a global-local depth regularization method. We observed that while depth scales may differ across frames and views, the relative depth order between pixel pairs should remain consistent. This insight led us to introduce a global ranking loss that distills temporally and spatially consistent depth relationships to guide 4DGS learning. At the local scale, we implement a patch-based smoothing constraint to address piece-wise discontinuities in the local geometries. Our approach guides the optimization by promoting global consistency across spacetime while minimizing localized artifacts, enhancing the coherent geometry and appearance learning within the 4D volume.

We evaluate GC-4DGS with state-of-the-art baselines on both the consumer GPU and edge device. Our method effectively outperforms various baselines and achieves higher rendering quality on the challenging N3DV \cite{li2022neural} and Technicolor \cite{sabater2017dataset} datasets (see Fig. \ref{fig.1}) with easy deployability on the edge device. Our contributions can be highlighted as follows:

\begin{itemize}
    \item To the best of our knowledge, this is among the first work to systematically tackle sparse-input dynamic Gaussian Splatting under multi-view settings, a challenging issue that widely exists in practical AIoT applications.
\end{itemize}

\begin{itemize}
    \item We exploit geometry priors from MVS to address the under-constrained optimization problem. A dynamic consistency checking strategy is introduced to further reduce estimation uncertainties of MVS across spacetime.  
\end{itemize}

% \begin{itemize}
%     \item We propose a global-local depth regularization method to ensure temporally consistent depth ranking while maintaining local smoothness. An adaptive Step Decay strategy is proposed to improve optimization stability.
% \end{itemize}

\begin{itemize}
    \item We propose a global-local depth regularization method to ensure consistent 4D geometry ranking while maintaining local smoothness, enhancing the spatiotemporal consistent reconstruction of both geometry and appearance.
\end{itemize}

% \begin{itemize}
%     \item We conduct experiments on GC-4DGS, which achieves state-of-the-art rendering quality in standard multi-view datasets, without sacrificing training and rendering efficiency.
% \end{itemize}

\begin{itemize}
    \item We demonstrate that GC-4DGS achieves state-of-the-art rendering quality across standard datasets and is easily deployable on resource-constrained IoT edge devices.
\end{itemize}

\begin{figure*}[t]
  \centering
   \includegraphics[width=1.0\linewidth]{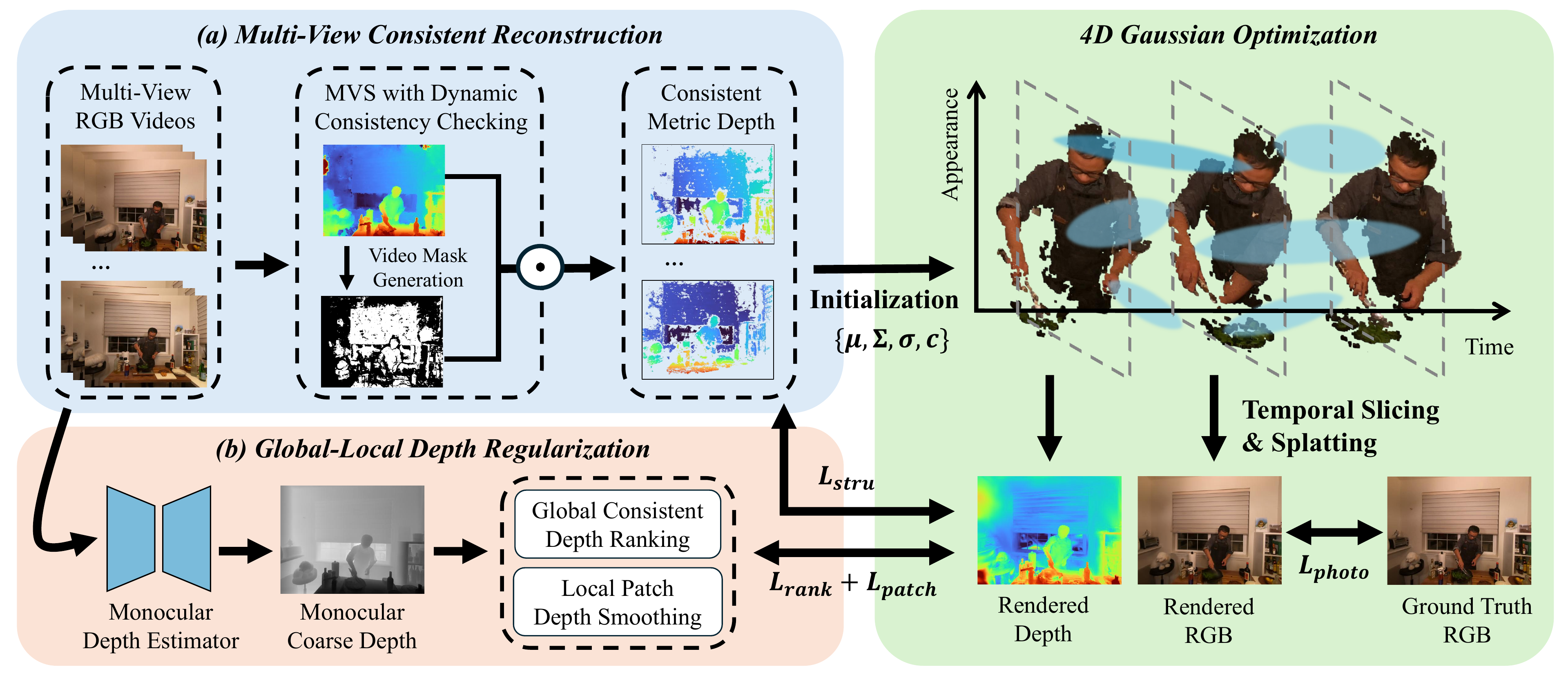}

   \caption{Framework Overview. (a) We introduce a dynamic consistency checking strategy to fuse view-consistent metric depths from a learning-based MVS method, which is then employed to obtain point clouds for Gaussian initialization and to supervise the learning of 4D geometry. (b) We propose a global-local depth regularization method to distill robust geometry information from a pre-trained MDE, which ensures consistent depth ranking while maintaining local patch smoothness. The optimization is conducted through temporal slicing, differentiable rendering, and color and depth supervision.}
   \label{fig.2}
\end{figure*}

\section{Related Work}

\subsection{Dynamic Gaussian Splatting}

The recently emerging Gaussian Splatting~\cite{kerbl20233d} has reshaped the landscape of dynamic radiance fields due to its efficiency and flexibility. The pioneering work by Luiten et al. \cite{luiten2024dynamic} enhances Gaussian Splatting with per-frame motions. Similarly, STG \cite{li2024spacetime} model dynamic motions of Gaussian points using polynomial functions. A prevalent approach, including E-D3DGS \cite{bae2025per}, involves decoupling dynamic scenes into a canonical space and a deformation field. Other research such as 4DGaussians \cite{wu20244d} employs decomposed 4D neural voxels for estimating Gaussian deformations. Concurrently, 4DGS \cite{yang2024real, duan20244d} lifts Gaussian Splatting into higher dimensions by designing spatiotemporal coherent 4D Gaussians. Unlike the aforementioned methods, 4DGS preserves the explicit nature of Gaussian primitives and utilizes a time-slicing operation that enables the capturing of complex dynamics, including abrupt appearances and disappearances. 

While existing dynamic Gaussian Splatting models achieve remarkable rendering quality, they are prone to overfitting when trained with a limited number of input views, which hinders their real-world applications in AIoT systems.

\subsection{Sparse-Input Dynamic View Synthesis}

Synthesizing novel views of dynamic scenes is difficult as the intrinsic correlations across timestamps should be captured effectively. To achieve high-fidelity DVS, most existing methods rely on dense input views obtained from either handheld cameras \cite{yang2024deformable} or complex camera array systems \cite{li2024spacetime}. To mitigate the dependency on dense multi-view videos, Shuai et al. \cite{shuai2022novel} design a layered neural representation that separates moving objects from static backgrounds. Similarly, Li et al. \cite{li2023dynamic} incorporate a spatiotemporal feature warping mechanism into a deep neural network to enable end-to-end DVS with sparse input views.  In addition, RF-DeRF \cite{somraj2024factorized} incorporates motion priors to guide the learning of the dynamic radiance field, while MoDGS \cite{qingming2025modgs} develop a pipeline to achieve DVS from static or slow-moving monocular videos. In contrast, our work focuses on multi-view settings and facilitates fast, high-quality rendering of dynamic scenes from sparse input views. 

\subsection{Geometry-Guided Gaussian Splatting}

Existing per-scene geometry-guided Gaussian Splatting methods can be categorized into three groups. Several studies \cite{li2024dngaussian,zhu2025fsgs,paliwal2025coherentgs} incorporates depth regularization derived from pre-trained MDEs to enhance sparse-view reconstruction. Other research \cite{chen2025mvsplat, charatan2024pixelsplat} investigates feed-forward 3DGS models by leveraging geometrical cues. Additionally, several works explore geometry information from 3D foundation models \cite{fan2024instantsplat} or MVS \cite{xu2024mvpgs,liu2024mvsgaussian} to improve the rendering quality. Although these solutions show impressive performance, they primarily focus on few-shot view synthesis for static scenes. To our knowledge, sparse-input DVS using Gaussian representations has been scarcely investigated.

\section{Preliminary}\label{sec 3}
4D Gaussian Splatting~\cite{yang2024real,duan20244d} represents the underlying volume of dynamic scenes using a set of anisotropic 4D Gaussian ellipsoids. Each Gaussian primitive can be described with a 4D center ${\boldsymbol{\mu}}={\left( {{\mu _x},{\mu _y},{\mu _z},{\mu _t}} \right)} \in {{\mathbb{R}}^{4}} $, a 4D covariance matrix ${\Sigma}\in {{\mathbb{R}}^{4\times4}}$, opacity $\sigma \in [0,1]$, and color $c$ represented by 4D spherical harmonics (SH). Similar to ~\cite{kerbl20233d}, the ${\Sigma}$ can be decomposed with 4D rotation matrix $\boldsymbol{\mathbf{R}} \in$ SO(4) and 4D scaling factors $\boldsymbol{\mathbf{S}} = \mathrm{diag}(s_x,s_y,s_z,s_t) \in {{\mathbb{R}}^{4}}$ as:
\begin{align}
{{\Sigma} = {\boldsymbol{\mathbf{R}}}{\boldsymbol{\mathbf{S}}}\boldsymbol{\mathbf{S}}^T\boldsymbol{\mathbf{R}}^T = \left( {\begin{array}{*{20}{c}}
\boldsymbol{\mathbf{A}}&\boldsymbol{\mathbf{M}}\\
{{\boldsymbol{\mathbf{M}}^T}}&\boldsymbol{\mathbf{Z}}
\end{array}} \right)}
\end{align}
where ${\boldsymbol{\mathbf{A}}}$ is a $3 \times 3$ matrix. When rendering the dynamic scene at time $t$, all 4D Gaussians are sliced into 3D space using the properties of multivariate Gaussian distribution:
\begin{align}
{{\mathcal{G}_{3D}}\left( {\boldsymbol{x},t} \right) = {\mathcal{F}}\left( t \right){e^{ - \frac{1}{2}{{\left[ {\boldsymbol{x} - {\boldsymbol{\mu} _{3D}}\left( t \right)} \right]}^T}\Sigma _{3D}^{ - 1}\left[ {\boldsymbol{x} - {\boldsymbol{\mu} _{3D}}\left( t \right)} \right]}}}
\end{align}
where ${\mathcal{F}}\left( t \right)$ denotes a temporal decay term, $\boldsymbol{\mu} _{3D}$ and $\Sigma _{3D}$ are the 3D center and covariance matrix of the sliced 3D Gaussians, respectively. When rendering, following the fast differentiable rasterization\cite{kerbl20233d}, the color of pixels on the 2D image at time $t$ is obtained as:
% \begin{equation}
% \begin{aligned}
% \label{eq3}
%     & {\boldsymbol{\mu} _{3D}}\left( t \right) = {\left( {{\mu _x},{\mu _y},{\mu _{\rm{z}}}} \right)^T} + \left( {t - {\mu _t}} \right)\boldsymbol{\mathbf{M}}/\boldsymbol{\mathbf{Z}}, \\
%     & {\Sigma _{3D}} = \boldsymbol{\mathbf{A}} - \boldsymbol{\mathbf{M}}{\boldsymbol{\mathbf{M}}^T}/\boldsymbol{\mathbf{Z}}.
% \end{aligned}
% \end{equation}

% During optimization, 4DGS adaptively controls the Gaussian densification by cloning and splitting 4D Gaussians with large spatiotemporal positional gradients \cite{yang2024real}. 

% When rendering, following the fast differentiable rasterization\cite{kerbl20233d}, the color of each pixel on the 2D image at time $t$ is obtained as:
\begin{align}
\label{eq4}
{{\mathcal{C}}\left( t \right) = \sum\limits_{i = 1}^N {{c_i}\left( t \right){\alpha _i}\left( t \right)\prod\limits_{j = 1}^{i - 1} {\left( {1 - {\alpha _j}\left( t \right)} \right)} } }
\end{align}
where ${\alpha _i}\left( t \right)$ is the projected opacity of $i$-th sliced 3D Gaussian at time $t$, and ${c_i}\left( t \right)$ is the RGB color after evaluating SH coefficients with view direction and timestamp. 

\section{Method}\label{sec 4}
Given a sparse set of synchronized videos $V^{t,n}$ (as few as 3 views) with known camera poses, where $t\in\{1, 2,..., N_f\}$ denotes the temporal index among $N_f$ frames and $n\in\{1, 2,..., N_c\}$ denotes the camera index among $N_c$ viewpoints, our objective is to synthesis photorealistic novel views for arbitrary timestamps and viewpoints within the spatiotemporal domain. Our framework, which builds upon 4DGS \cite{yang2024real} (Sec. \ref{sec 3}), enables fast, high-fidelity reconstruction of dynamic scenes from sparse RGB videos. To reduce the estimation uncertainties in MVS, we introduce a dynamic consistency checking strategy to remain more reliable and accurate depth values from MVS (Sec. \ref{sec 4.1}). Additionally, we propose a global-local depth regularization method (Sec. \ref{sec 4.2}) to guide the learning of 4D geometry. An overview of the framework is illustrated in Fig. \ref{fig.2}. 

\subsection{Multi-View Consistent Reconstruction}\label{sec 4.1}
To address the ill-posed geometric optimization problem, we utilize MVSFormer~\cite{cao2022mvsformer}, a recent learning-based MVS method with impressive generalization abilities for both geometry initialization and subsequent structural supervision. With sparse-view video streams $V^{t,n}$ as input, each view is treated as a reference view, with the remaining serving as neighbor views. All training videos are sliced by each time instant and the pre-trained MVS network is employed to construct a cost volume and predict pixel-wise metric depth streams $D_{mvs}^{t,n}$.

\vspace{0.5em}
\noindent\textbf{Dynamic Consistency Checking.} The generalizable MVS suffers from several inherent limitations, such as significant noise and inaccurate estimations with low visual overlap. Inspired by ~\cite{yan2020dense}, we introduce a dynamic consistency checking strategy to mitigate these issues. Specifically, for a pixel $\boldsymbol{p}$ in the reference view $n$ at time $t$ with predicted depth $D^{t,n}(\boldsymbol{p})$, we project it into one of the neighbor views at the same timestamp to obtain pixel $\boldsymbol{p}'$ and its depth value. Then we re-project $\boldsymbol{p}'$ back to the reference view as $\boldsymbol{p}''$ and determine its depth $D^{t,n}(\boldsymbol{p}'')$. The re-projection errors are calculated as:

\begin{equation}
\begin{aligned}
& {\xi_p^{t,n} = \|{\boldsymbol{p} - \boldsymbol{p}''}\|_2}, \\
& {\xi_d^{t,n}=\frac{\|{D^{t,n}(\boldsymbol{p})-D^{t,n}(\boldsymbol{p}'')}\|_1}{D^{t,n}(\boldsymbol{p})}}.
\end{aligned}
\end{equation}
where $\xi_p^{t,n}$ and $\xi_d^{t,n}$ represent the back-projection errors in pixel space and depth space, respectively. To achieve dynamic aggregation of geometric matching errors across spacetime, we define the geometric consistency among viewpoints and frames as:
\begin{align}
c(\boldsymbol{p})=  \frac{1}{N_f} \sum\limits_{t = 1}^{N_f} \sum\limits_{n = 1}^{N_c}{e^{-(\xi_p^{t,n}+\beta\xi_d^{t,n})}}
\end{align}
where $\beta$ is a parameter that balances the two re-projection metrics. We calculate the dynamic geometric consistency for each pixel and filter out unreliable pixels where $c(\boldsymbol{p})$ falls below a predefined threshold. This process produces the geometric consistency masks for the raw depth streams estimated by MVS. By integrating the masks with photometric probability maps derived from the MVS model \cite{cao2022mvsformer}, we can obtain the dynamic consistency masks $M^{t,n}$, which can be used to fuse multi-view metric depth for 4D Gaussian initialization, as illustrated in Fig. \ref{fig.2} (a).

\vspace{0.5em}
\noindent\textbf{4D Structure Supervision.}
The metric depth streams from MVS can provide reliable 4D structure supervision for the optimization of Gaussian primitives. We first implement a differentiable depth rasterizer by using the alpha-blending aggregation approach described in Eq. (\ref{eq4}) to render pixel-wise depth values at specific timestamps:
\begin{align}
{{\mathcal{D}}\left( t \right) = \sum\limits_{i = 1}^N {{d_i}\left( t \right){\alpha _i}\left( t \right)\prod\limits_{j = 1}^{i - 1} {\left( {1 - {\alpha _j}\left( t \right)} \right)} } }
\end{align}
where $d_i$ denotes the z-buffer of $i$-th sliced 3D Gaussian at time $t$, and ${\alpha _i}$ is identical to that in Eq. (\ref{eq4}).

To ensure the multi-view consistent structure during Gaussian optimization, we add constraints on the rendered depths with high-confidence MVS depths:
\begin{align}
{\mathcal{L}_{stru} = \sum{\text{smooth}}_{L_1}(({D}_{ren}-{{D}_{mvs}})\odot M) }
\end{align}
in which $\text{smooth}_{L_1}$ is a robust loss function to facilitate convergence around the zero point \cite{girshick2015fast}, ${D}_{ren}$ is the rendered depth map, ${D}_{mvs}$ is the estimated metric depth from MVS, and $M$ is the mask obtained from dynamic consistency checking.

\subsection{Global-Local Depth Regularization}\label{sec 4.2}
The metric depth may lack guidance in under-reconstructed regions. In areas where MVS supervision is insufficient, we use monocular depth priors to provide compensation. Previous depth-supervised Gaussian radiance fields typically construct depth loss based on either the source scales \cite{zhu2025fsgs} or normalized scales \cite{li2024dngaussian,paliwal2025coherentgs} of depth maps. Despite various alignment techniques proposed to address scale ambiguity, depths from MDEs lack spatiotemporal correlations. Specifically, the depth ratios between the same pixel pairs can vary across frames and views, even after normalization \cite{qingming2025modgs}. Consequently, these methods may result in scale inconsistency when modeling the underlying 4D geometry of dynamic scenes.

\begin{figure}[t]
  \centering
   \includegraphics[width=1.0\linewidth]{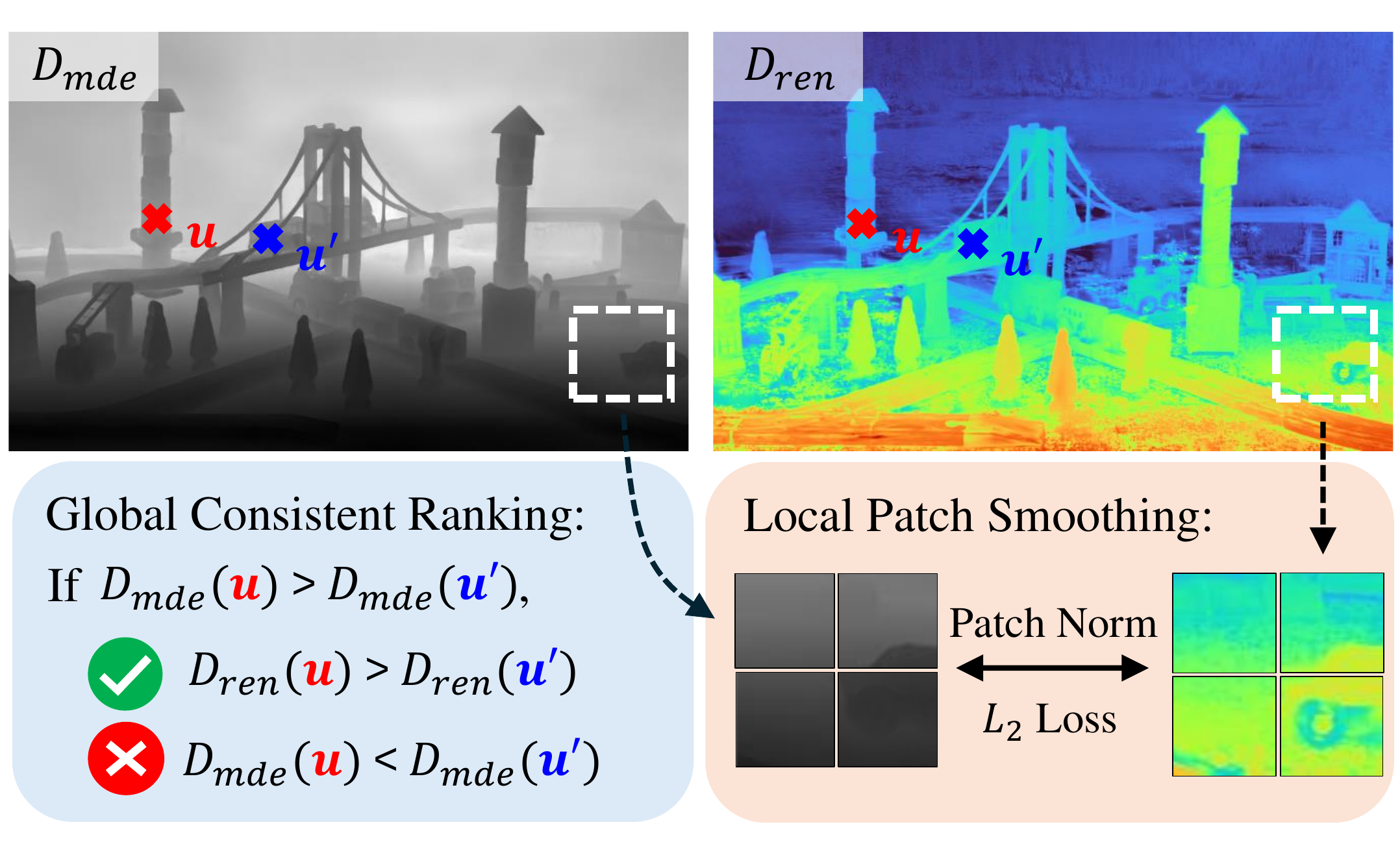}
   \caption{Global-Local Depth Regularization. For a randomly selected pixel pair $(\boldsymbol{u}, \boldsymbol{u'})$, we enforce that the relative order in $D_{ren}$ is consistent with that in $D_{mde}$. Additionally, we encourage local smoothness by applying absolute supervision on normalized depth patches. These methods enable our model to distill robust geometric information from the monocular depth maps, thereby facilitating the learning of 4D geometry.}
   \label{fig.3}
\end{figure}
\vspace{0.5em}
\noindent\textbf{Global Consistent Depth Ranking.}
We observed that while MDE-derived depth scales may differ across frames and views, the relative depth order between pixel pairs from $D_{ren}$ and $D_{mde}$ should remain consistent, as illustrated in Fig. \ref{fig.3}. Inspired by \cite{wang2023sparsenerf,qingming2025modgs}, we propose a global ranking loss to regularize the ordinal relationship between sampled point pairs. Drawing ideas from the nonlinear transformation in neural networks, we use $Sigmoid(\cdot)$ to push the depth ranking of the selected pixel pair $(\boldsymbol{u}, \boldsymbol{u}')$ in $D_{ren}$ towards $0$ or $1$:
\begin{align}
{{\mathcal{R}}_{ren}^{t,n} = Sigmoid(\kappa(D_{ren}^{t,n}(\boldsymbol{u})-D_{ren}^{t,n}(\boldsymbol{u}'))) }
\end{align}
where $t$ and $n$ represent a specific timestamp and camera view, $\kappa$ is a positive constant that maps the depth difference to $+\infty$ or $-\infty$, and $D_{ren}(\boldsymbol{u})$ means the rendered depth value on the pixel $\boldsymbol{u}$. We then define an indicator function for $D_{mde}$ as the ground truth ranking:
\begin{align}
{{\mathcal{R}}_{mde}^{t,n} = \begin{cases} 
0, & \text{if } D_{mde}^{t,n}(\boldsymbol{u}) < D_{mde}^{t,n}(\boldsymbol{u}') \\
1, & \text{if } D_{mde}^{t,n}(\boldsymbol{u}) > D_{mde}^{t,n}(\boldsymbol{u}') 
\end{cases} }
\end{align}
where ${\mathcal{R}}_{mde}$ is the ranking indicator function on depth maps from $D_{mde}$, and $D_{mde}(\boldsymbol{u})$ means the depth value from MDEs on the pixel $\boldsymbol{u}$. Finally, the global consistent ranking loss is formulated as:
\begin{align}
{{\mathcal{L}}_{rank} = \sum_{(\boldsymbol{u}, \boldsymbol{u}')}\|{\mathcal{R}}_{ren}^{t,n}-{{\mathcal{R}}_{mde}^{t,n}}\| }
\end{align}
In our study, we select a fixed number of pixel pairs from the full video resolution at each training iteration to calculate the ranking loss, making our method more robust to scale-inconsistent depth priors derived from pre-trained MDEs.

\vspace{0.5em}
\noindent\textbf{Local Patch Depth Smoothing.}
While the ranking constraint ensures the global consistency of the 4D geometry, it cannot guarantee the local continuity of the rendered depth. To fill the gap of global supervision, we encourage the local patch smoothness for depth regularization \cite{li2024dngaussian}. Instead of using both rendered depths and estimated depths in the source scale, we divide the entire depth map into small patches $\boldsymbol{e}$ and perform local normalization on each pixel of the patch:
\begin{align}
{\hat{D}(\boldsymbol{p}) = \frac{{D}(\boldsymbol{p})-\text{mean}(D(\boldsymbol{e}))}{\text{std}(D(\boldsymbol{e})) + \delta}},   \boldsymbol{p} \in {\boldsymbol{e}}
\end{align}
where $\delta$ is a value to avoid numerical instability. After that, we impose $L_2$ loss term on all patches from both normalized rendered depths and MDE depths, respectively:

\begin{align}
{{\mathcal{L}}_{patch} =  \sum_{\boldsymbol{e}}\|{\hat{D}_{ren}^{t,n}(\boldsymbol{e})-{\hat{D}_{mde}^{t,n}(\boldsymbol{e})}}\| }
\end{align}
In practice, we sample a patch size from a pre-defined range based on the geometric complexity of scenes, i.e., their global depth ranges and local depth changes, and $\delta = 2\times10^{-4}$.
% Finally, the overall loss function for depth regularization is defined as:
% \begin{align}
% {{\mathcal{L}}_{depth} = \lambda_1{\mathcal{L}}_{rank}+\lambda_2{\mathcal{L}}_{patch}+\lambda_3{\mathcal{L}}_{stru}}
% \end{align}

\subsection{Optimization}\label{sec 4.3}
% \noindent\textbf{Step Decay Densification.} Unlike the fixed densification threshold used in \cite{yang2024real}, we adopt an adaptive step decay strategy to improve optimization stability in sparse-view settings. Specifically, the threshold decays from $6 \times 10^{-4}$ to $2 \times 10^{-4}$ with a factor of 0.8 every 1000 iterations. This approach effectively mitigates training instability, accelerates convergence, and improves overall training efficiency.

% \vspace{0.5em}
\noindent\textbf{Objective Function.} The total loss function of our approach is formulated by:
\begin{align}
\label{eq15}
{{\mathcal{L}}_{total} ={\mathcal{L}}_{photo}+\lambda_1{\mathcal{L}}_{rank}+\lambda_2{\mathcal{L}}_{patch}+\lambda_3{\mathcal{L}}_{stru}}
\end{align}
where $\lambda_1$, $\lambda_2$ and $\lambda_3$ are hyper-parameters. The photometric loss ${\mathcal{L}}_{photo}$ is calculated by comparing the rendered image at different time steps with the ground truth images using a combination of $L_1$ loss and D-SSIM loss, as proposed in \cite{kerbl20233d}. 

% we adopt a Step Decayed strategy to enhance optimization stability in sparse-view settings. Specifically, the threshold decays from $6 \times 10^{-4}$ to $2 \times 10^{-4}$ with a factor of 0.8 every 1000 iterations. This approach effectively mitigates training instability, accelerates convergence, and improves overall training efficiency.

% \begin{figure}[t]
%   \centering
%    \includegraphics[width=1.0\linewidth]{Figure/figure4v4.pdf}
%    \caption{RGB and geometry visualization of baselines and GC-4DGS on N3DV Dataset with 3 input views.}
%    \label{fig.5}
% \end{figure}
% % \setlength{\textfloatsep}{6pt}     % Adjust this value as needed

\begin{table}[t]
    \centering
    \caption{Quantitative analysis on N3DV Dataset with 3 input views. *: STG is tested using the first 50 frames. FPS and training time (hours) are measured at $1352 \times 1014$ resolution.}
    \label{tab1}
    \setlength{\tabcolsep}{2.5pt}
    \begin{tabular}{l|cccc|cc}
        \toprule
        Methods & PSNR$\uparrow$ & SSIM$\uparrow$ & LPIPS$\downarrow$ & AVGE$\downarrow$ & FPS$\uparrow$ & Train$\downarrow$ \\
        \midrule
        HexPlane\cite{cao2023hexplane}    & 15.54 & 0.493 & 0.499 & 0.215 & -     & - \\
        MixVoxels\cite{wang2023mixed}  & 15.82 & 0.724 & 0.499 & 0.190 & 16    & 1.3h\\
        HyperReel\cite{attal2023hyperreel}  & 18.90 & 0.708 & 0.334 & 0.132 & 2.2   & 4.1h \\
        K-Planes\cite{fridovich2023k}   & 23.65 & 0.831 & 0.247 & 0.076 & 0.6   & 2.6h\\
        RF-DeRF\cite{somraj2024factorized}  & 25.07 & 0.878 & 0.215 & 0.062 & -      & -\\
        \midrule
        STG$^{*}$\cite{li2024st}       & 25.76 & \cellcolor{yellow!50}0.894 & 0.094 & \cellcolor{yellow!50}0.043 & 322    & 0.5h\\
        4DGaussians\cite{wu20244d} & 25.40 & 0.885 & 0.096 & 0.046 & 34    & 0.7h\\
        4DGS\cite{yang2024real}       & \cellcolor{yellow!50}26.11 & 0.882 & 0.110 & 0.045 & 183  & 2.8h\\
        E-D3DGS\cite{bae2025per}    & 25.75 & 0.875 & \cellcolor{yellow!50}0.088 & 0.044 & 177  & 0.8h \\
        GC-4DGS        & \cellcolor{red!50}\textbf{27.69} & \cellcolor{red!50}\textbf{0.907} & \cellcolor{red!50}\textbf{0.074} & \cellcolor{red!50}\textbf{0.034} & 190    & 1.1h\\
        \bottomrule
    \end{tabular}
    \label{tab1}
\end{table}

\begin{table}[t]
    \centering
    \setlength{\tabcolsep}{2.5pt}
    \caption{Quantitative analysis on Technicolor Dataset with 3 input views. ${\dagger}$: Results are tested using 300 frames. FPS and training time (hours) are measured at $1024 \times 544$ resolution.}
    \begin{tabular}{l|cccc|cc}
        \toprule
        Methods & PSNR$\uparrow$ & SSIM$\uparrow$ & LPIPS$\downarrow$ & AVGE$\downarrow$ & FPS$\uparrow$ & Train$\downarrow$\\
        \midrule
        HexPlane$^{\dagger}$\cite{cao2023hexplane}    & 13.86 & 0.267 & 0.542 & 0.267 & - & -\\
        HyperReel\cite{attal2023hyperreel}  & 18.78 & 0.494 & 0.374& 0.152 & 7.3 & 4.4h \\
        K-Planes$^{\dagger}$\cite{fridovich2023k}   & 18.75 & 0.665 & 0.305 & 0.133 & - & -\\
        RF-DeRF$^{\dagger}$\cite{somraj2024factorized}    & 20.41 & 0.742 & 0.251 & 0.105 & - & -\\
        \midrule
        STG\cite{li2024st}       & \cellcolor{yellow!50}23.23 &  \cellcolor{yellow!50}0.778 & \cellcolor{red!50}\textbf{0.127} & \cellcolor{yellow!50}0.066 & 226      & 0.4h\\
        4DGaussians\cite{wu20244d} & 19.01 & 0.545 & 0.299 & 0.136 & 67    & 0.6h\\
        4DGS\cite{yang2024real} & 21.78 & 0.736 & 0.258 & 0.096 & 121   & 3.3h \\
        E-D3DGS\cite{bae2025per}  & 21.96 & 0.718 & 0.199 & 0.088 & 46    & 4.2h\\
        GC-4DGS        & \cellcolor{red!50}\textbf{23.96} & \cellcolor{red!50}\textbf{0.787} & \cellcolor{yellow!50}0.144 & \cellcolor{red!50}\textbf{0.064} & 115          & 1.0h\\
        \bottomrule
    \end{tabular}
    \label{tab2}
\end{table}

\section{Experiments}\label{sec 4}

\begin{figure*}[t]
  \centering
   \includegraphics[width=1.0\linewidth]{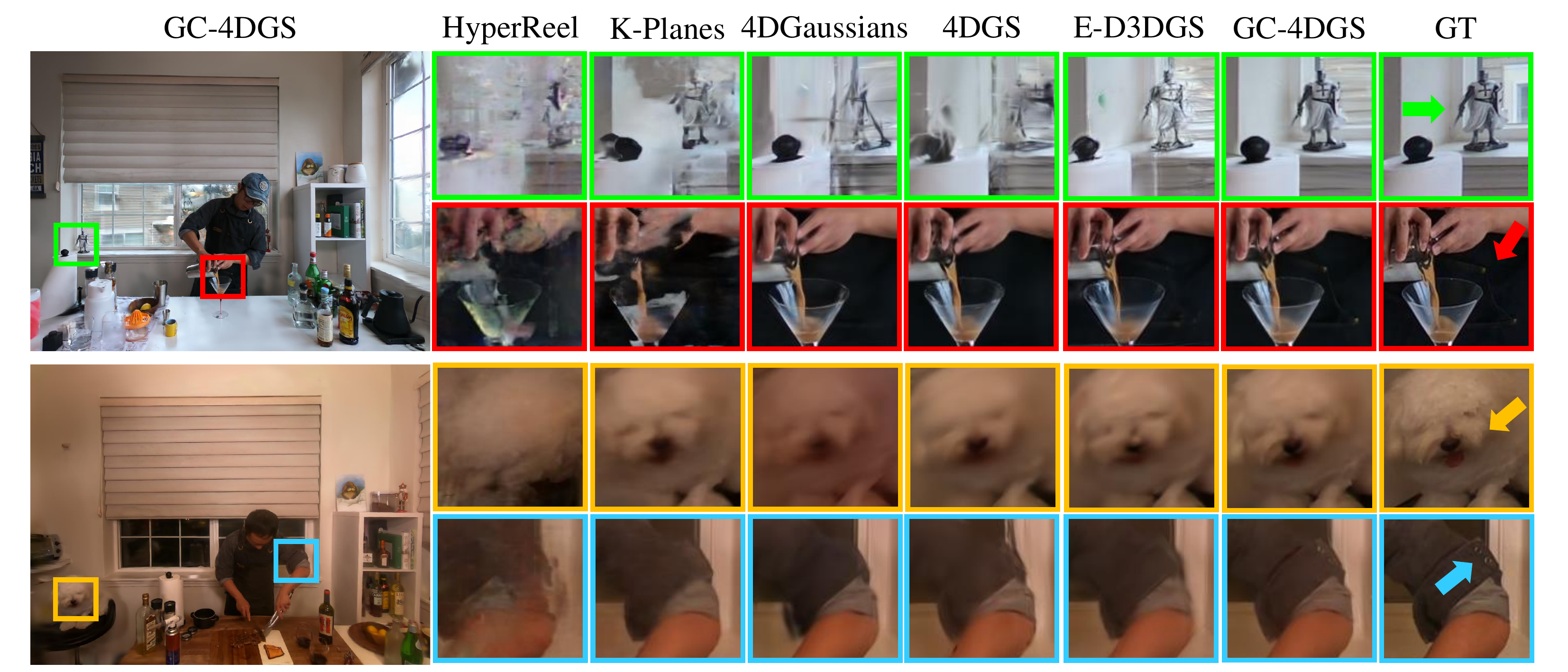}

   \caption{Qualitative results on N3DV Dataset with 3 input views. GC-4DGS achieves consistent improvement in both static and dynamic regions when compared to other state-of-the-art dynamic radiance fields.}
   \label{fig.4}
\end{figure*}

\begin{figure*}[t]
  \centering
   \includegraphics[width=1.0\linewidth]{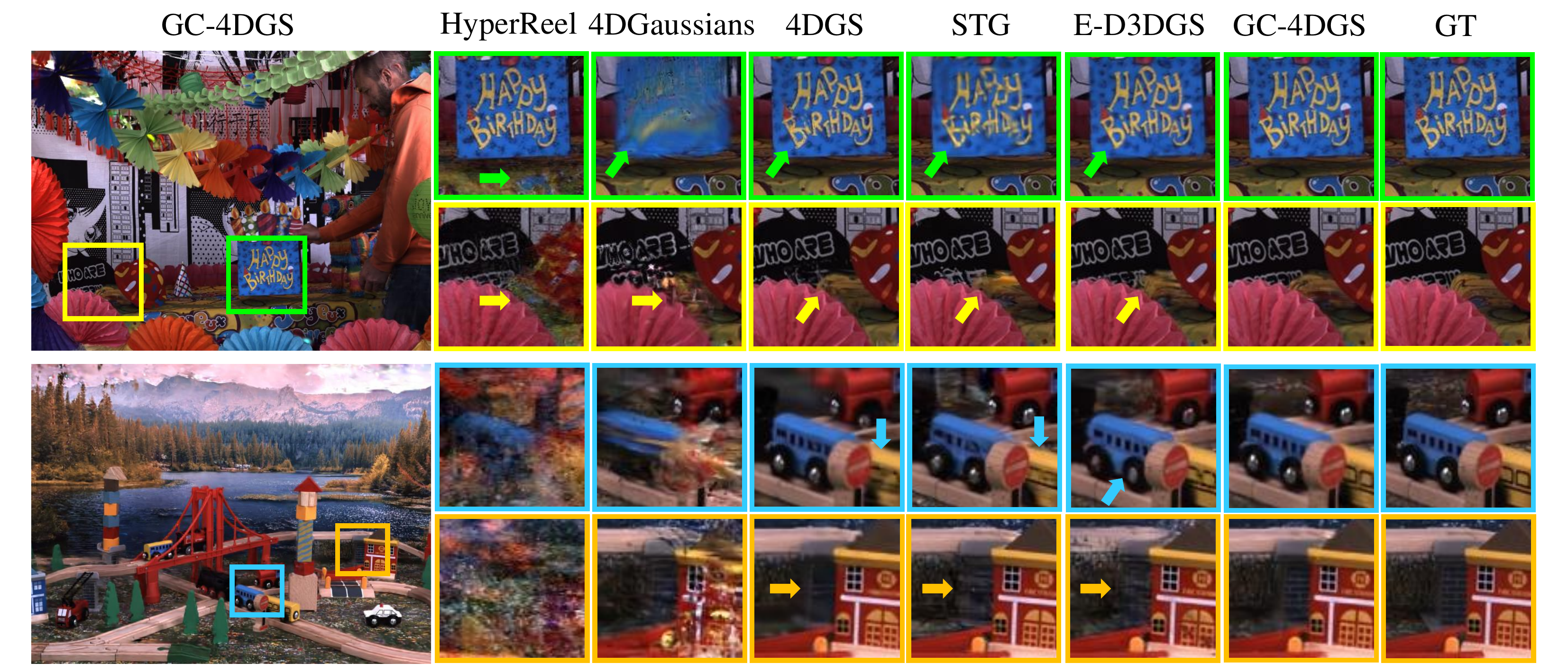}
   \caption{Qualitative results on Technicolor Dataset with 3 input views. HyperReel \protect\cite{attal2023hyperreel} and 4DGaussians \protect\cite{wu20244d} produce significantly distorted images, while STG \protect\cite{li2024spacetime} struggles to capture complex dynamics. 4DGS \protect\cite{yang2024real} and E-D3DGS \protect\cite{bae2025per} are prone to overfit in areas with limited observations and produce blurred results.}
   \label{fig.6}
\end{figure*}

\subsection{Datasets and Evaluations}
We conduct experiments on two popular multi-view dynamic scene benchmarks: N3DV Dataset \cite{li2022neural} and Technicolor Dataset \cite{sabater2017dataset}. Following previous settings, we downsample the recorded videos in half resolution for all the experiments and select the same training and testing views as in RF-DeRF \cite{somraj2024factorized}. 

\vspace{0.5em}
\noindent\textbf{N3DV Dataset.} N3DV Dataset \cite{li2022neural} contains six 10-second indoor video sequences captured by 18 to 21 synchronized cameras at 30 FPS. We utilize the center camera for evaluation purposes, while uniformly selecting 3 input videos from the remaining cameras to serve as training views. We use 300 frames from each scene for training.

\vspace{0.5em}
\noindent\textbf{Technicolor Dataset.} Technicolor Dataset \cite{sabater2017dataset} consists of videos captured using a $4 \times 4$ camera array. In alignment with RF-DeRF \cite{somraj2024factorized}, we train baseline models on five commonly used scenes (i.e., Birthday, Painter, Remy, Theater, and Train) with 3 input views, reserving the center video for testing. We select 50 frames from each scene for our experiments.

\vspace{0.5em}
\noindent\textbf{Metrics.} We evaluate the quality of rendered images using Peak Signal-to-Noise Ratio (PSNR), Structural Similarity Index (SSIM), and Learned Perceptual Image Patch Similarity (LPIPS) with an AlexNet backbone. To ease comparison, we also report the Average Error (AVGE), the geometric mean of MSE=$10^{-\mathrm{PSNR}/10}$, $\sqrt{1-\mathrm{SSIM}}$, and LPIPS \cite{li2024dngaussian}. Additionally, we use FPS to evaluate the rendering speed.

\vspace{0.5em}
\noindent\textbf{Implementation Details.} 
Following \cite{yang2024real}, we implement our method using Pytorch and CUDA. To generate geometry-consistent depth maps, we employ the pre-trained MVSFormer \cite{cao2022mvsformer}. The dynamic consistency checking is conducted with $\beta=200$ and a threshold of 1.8. We use only the depth maps from the first frame for point cloud fusion. Additionally, we utilize DepthAnythingV2 \cite{yang2024depth} to estimate monocular depth streams. We randomly select 500k pixel pairs to calculate the ranking loss at each iteration. The range of patch size is set as [8, 16] and [16, 32] for N3DV and Technicolor datasets, respectively. We set $\lambda_1 = 0.05$, $\lambda_2 = 0.02$ and $\lambda_3 = 0.02$ in Eq. (\ref{eq15}). The primary experiments for each scene are performed on a single RTX 4090 GPU. 

\subsection{Results Analysis}
We compare GC-4DGS against state-of-the-art dynamic radiance fields, including HexPlane \cite{cao2023hexplane}, HyperReel \cite{attal2023hyperreel}, MixVoxels \cite{wang2023mixed}, K-Planes \cite{fridovich2023k}, RF-DeRF \cite{somraj2024factorized}, STG \cite{li2024spacetime}, 4DGaussians \cite{wu20244d}, 4DGS \cite{yang2024real} and E-D3DGS \cite{bae2025per}.

\vspace{0.5em}
% \noindent\textbf{Comparisons on N3DV Dataset.} As shown in Tab. \ref{tab1}, GC-4DGS outperforms all existing dynamic radiance field-based approaches on rendering quality while maintaining a fast rendering speed. GC-4DGS surpasses RF-DeRF \cite{somraj2024factorized}, which is tailored for sparse-input DVS tasks, and 4DGS \cite{yang2024real} by 2.62dB and 1.58dB in PSNR, positioning our method as the top choice for sparse-view dynamic scenarios. Furthermore, GC-4DGS achieves a real-time rendering speed of 190 FPS, which exceeds that of dynamic NeRFs and deformation field-based Gaussian Splatting methods. Fig. \ref{fig.4} illustrates that the existing DVS methods face challenges with sparse inputs. In contrast, GC-4DGS recovers more textural and structural details in both central dynamic areas and under-observed static regions

\noindent\textbf{Comparisons on N3DV Dataset.} As shown in Tab. \ref{tab1}, GC-4DGS outperforms all existing dynamic radiance field-based approaches on rendering quality while maintaining a fast rendering speed. GC-4DGS surpasses RF-DeRF \cite{somraj2024factorized}, which is tailored for sparse-input DVS tasks, and 4DGS \cite{yang2024real} by 2.62dB and 1.58dB in PSNR, positioning our method as the top choice for sparse-view dynamic scenarios. Furthermore, GC-4DGS achieves a real-time rendering speed of 190 FPS, which exceeds that of dynamic NeRFs and deformation field-based Gaussian Splatting methods. Fig. \ref{fig.4} presents the qualitative comparison. HyperReel \cite{attal2023hyperreel} and K-planes \cite{fridovich2023k} fails to render the scene with few input views. Although Gaussian Splatting-based approaches improve rendering qualities, they struggle to capture fine-grained dynamics (e.g., the dog's hair) and under-observed geometry (e.g., the mini statue). In contrast, GC-4DGS recovers more textural and structural details in both central dynamic areas and under-observed static regions.

\noindent\textbf{Comparisons on Technicolor Dataset.} The results are presented in Tab. \ref{tab2} and Fig. \ref{fig.6}. GC-4DGS outperforms previous works across all rendering quality metrics except for LPIPS. Notably, STG \cite{li2024spacetime} is initialized using point clouds estimated from all timestamps, whereas we employ only the first frame for point cloud fusion, achieving a trade-off between complexity and performance during the initialization phase. 

\subsection{Ablation Studies}
\begin{figure}[t]
  \centering
   \includegraphics[width=1.0\linewidth]{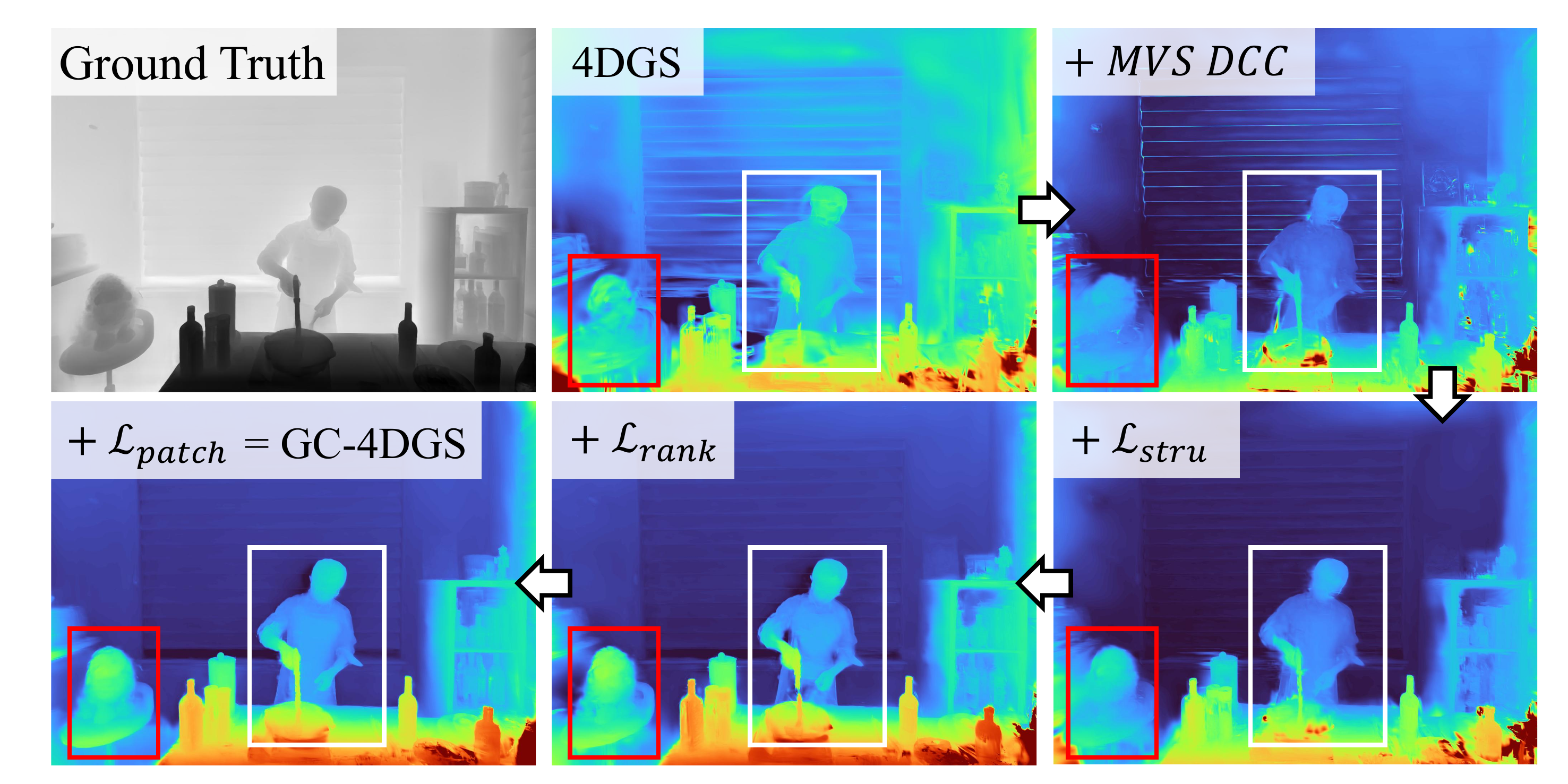}
   \caption{Ablation study on visual effects. We compare the depth of novel views to verify the effectiveness of components in GC-4DGS.}
   \label{fig.7}
\end{figure}

\noindent\textbf{Effectiveness of Initialization Strategies.} Tab. \ref{tab4} shows a significant performance improvement when equipping the baseline with point clouds initialized by MVS, especially through dynamic consistency checking (DCC). This is because MVS with dynamic consistency checking provides a richer and more accurate initialization compared to the vanilla COLMAP pipeline \cite{schoenberger2016sfm, schoenberger2016mvs} and MVS with downsampling (DS), thereby enhancing the subsequent optimization process. Fig. \ref{fig.7} illustrates that MVS contributes to the learning of fine-grained 4D geometry. However, the improvement is limited as the overfitting during sparse-view optimization persists.

\begin{table}
    \centering
    \caption{Ablation study on N3DV Dataset in 3-view settings. DS and DCC refer to downsampling and dynamic consistency checking, respectively.}
    \begin{tabular}{l|cccc}
        \toprule
        Method & PSNR$\uparrow$ & SSIM$\uparrow$ & LPIPS$\downarrow$ & AVGE$\downarrow$ \\
        \midrule
        Baseline                            & 26.11 & 0.882 & 0.110& 0.045 \\
        Baseline \textit{w/ MVS + DS}         & 26.72 & 0.890 & 0.095 & 0.041  \\
        Baseline \textit{w/ MVS + DCC}         & 26.85 & 0.893 & 0.089& 0.039 \\
        \midrule
        % GC-4DGS \textit{w/o SDD}       &  &  &  &  \\
        GC-4DGS \textit{w/o} ${\mathcal{L}}_{rank}$      & 27.17 & 0.898 & 0.084& 0.037\\
        GC-4DGS \textit{w/o} ${\mathcal{L}}_{patch}$     & 27.26 & 0.899 & \cellcolor{yellow!50}0.079& 0.036\\
        GC-4DGS \textit{w/o} ${\mathcal{L}}_{struc}$     & \cellcolor{yellow!50}27.41 & \cellcolor{yellow!50}0.904 & 0.081 & \cellcolor{yellow!50}0.036\\
        \midrule
        GC-4DGS                           & \cellcolor{red!50}\textbf{27.69} & \cellcolor{red!50}\textbf{0.907} & \cellcolor{red!50}\textbf{0.074}& \cellcolor{red!50}\textbf{0.034}\\
        \bottomrule
    \end{tabular}
    \label{tab4}
\end{table}
% \setlength{\textfloatsep}{6pt}     % Adjust this value as needed

% \begin{table}
%     \centering
%     % \setlength{\tabcolsep}{8pt}
%     \caption{Impact of Gaussian Initialization on N3DV Dataset.}
%     \begin{tabular}{l|cccc}
%         \toprule
%         Method & PSNR$\uparrow$ & SSIM$\uparrow$ & LPIPS$\downarrow$ & AVGE$\downarrow$\\
%         \midrule
%         Baseline                            & 26.11 & \cellcolor{yellow!50}0.882 & \cellcolor{yellow!50}0.110& \cellcolor{yellow!50}0.045 \\
%         Baseline \textit{w/ DUSt3R + DS}      & 19.52 & 0.732 & 0.546& 0.147 \\
%         Baseline \textit{w/ MASt3R + DS}      & \cellcolor{yellow!50}26.59 & 0.871 & 0.145& 0.049 \\
%         \midrule
%         Baseline \textit{w/ MVS + DCC}         & \cellcolor{red!50}\textbf{26.85} & \cellcolor{red!50}\textbf{0.893} & \cellcolor{red!50}\textbf{0.089} & \cellcolor{red!50}\textbf{0.039}\\
%         \bottomrule
%     \end{tabular}
%     \label{tab5}
% \end{table}
% \setlength{\textfloatsep}{6pt}     % Adjust this value as needed

% We also explore the use of state-of-the-art 3D foundation models, specifically DUSt3R \cite{wang2024dust3r} and MASt3R \cite{leroy2025grounding}, to achieve, as shown in Tab. \ref{tab5}. While these methods demonstrate effectiveness in dense and unconstrained 3D reconstruction tasks, they suffer from redundant point maps and suboptimal pose estimations \cite{fan2024instantsplat}. In contrast, our method performs better by producing high-confidence point clouds through the dynamic consistency checking method.

\vspace{0.5em}
\noindent\textbf{Effectiveness of Depth Regularization.} As presented in Tab. \ref{tab4} and illustrated in Fig. \ref{fig.7}, the proposed depth regularization components provide complementary benefits that enhance geometric consistency. Specifically, ${\mathcal{L}}_{struc}$ employs high-confidence metric depths to rectify geometric errors arising from the Gaussian representation. In regions not sufficiently covered by MVS, we incorporate ${\mathcal{L}}_{rank}$ and ${\mathcal{L}}_{patch}$ to harness robust geometric information from monocular depth priors. In particular, ${\mathcal{L}}_{rank}$ serves as the dominant term that enforces global consistency in depth ordering, while ${\mathcal{L}}_{patch}$ guarantees local smoothness to eliminate patch-level artifacts.
\vspace{0.5em}

% \noindent\textbf{Effectiveness of Step Decay Densification.}
% \vspace{0.5em}

\subsection{Deployability on IoT Edge Devices}\label{sec 5.4}
\begin{figure}[t]
  \centering
   \includegraphics[width=1.0\linewidth]{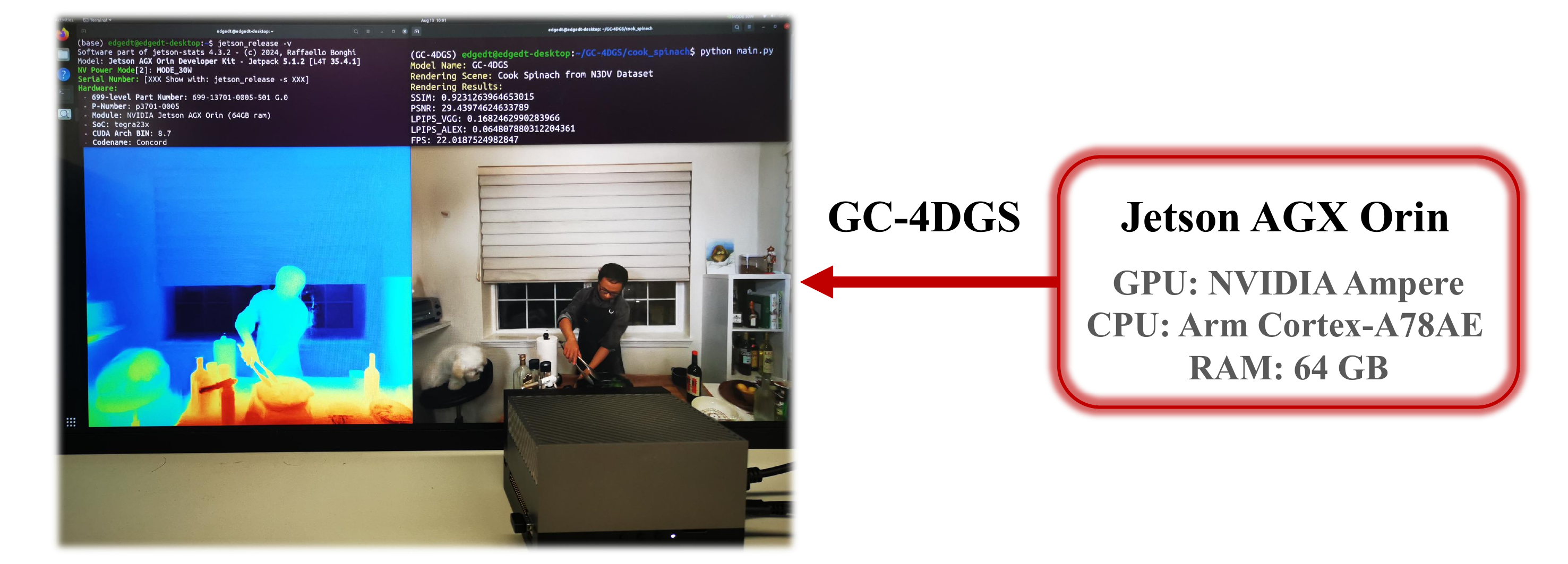}
   \caption{Deployment results of GC-4DGS on the IoT edge device. The
specifications of the testbed are shown in Sec. \ref{sec 5.4}.}
   \label{fig.8}
\end{figure}

To further evaluate the deployability of GC-4DGS on resource-constrained edge devices within IoT systems, experiments are conducted on the Jetson AGX Orin Developer Kit, which is equipped with a 1.3 GHz NVIDIA Ampere architecture GPU, a 2.2 GHz ARM Cortex-A78AE processor, and 64 GB of RAM. Our model was first trained in the cloud before being deployed on edge devices for testing. The results in Fig. \ref{fig.8} demonstrate that GC-4DGS is easily deployable on IoT edge devices with limited computing resource consumption, highlighting the potential of our model for AIoT applications.

\section*{Conclusion}
    In this paper, we present GC-4DGS, a novel framework for real-time dynamic view synthesis from sparse input views. A dynamic consistency checking method is introduced to reduce estimation uncertainties associated with data-driven MVS approaches. To address the spatiotemporal inconsistency inherent in depth priors estimated by MDEs, we propose a global-local depth regularization method that enforces global depth ranking consistency while preserving local patch smoothness. Experiments show that GC-4DGS achieves state-of-the-art rendering quality in sparse-input dynamic view synthesis without sacrificing the training and rendering efficiency. Future work will focus on improving the computational efficiency of GC-4DGS during training and rendering, especially for resource-constrained edge devices \cite{ye2025gaussian} and AIoT applications \cite{zhang20253dgstreaming}.
    
\section*{Acknowledgment}

This research was supported in part by the HK RGC Research Impact Fund (R5009-21), HK RGC Research Impact Fund (R5006-23), and the Research Institute for Artificial Intelligence of Things, The Hong Kong Polytechnic University.

\bibliographystyle{IEEEtran}
\bibliography{reference}
\end{document}